%% file: main.tex
\documentclass[10pt,twocolumn,letterpaper]{article}

\usepackage{cvpr}              
\usepackage[accsupp]{axessibility}
\usepackage{float}
\usepackage{dsfont}


\definecolor{cvprblue}{rgb}{0.21,0.49,0.74}
\usepackage[pagebackref,breaklinks,colorlinks,allcolors=cvprblue]{hyperref}

\def\paperID{12} 
\def\confName{CVPR}
\def\confYear{2025}

\title{IGL-DT: Iterative Global-Local Feature Learning with Dual-Teacher Semantic Segmentation Framework under Limited Annotation Scheme}

\author{Dinh Dai Quan Tran$^{1*}$ \quad Hoang-Thien Nguyen$^{2*}$ \quad  Thanh-Huy Nguyen$^{3}$ \\ Gia-Van To$^{4}$  \quad
Tien-Huy Nguyen$^{5}$ \quad Quan Nguyen$^{6}$ \\ \\
$^{1}$National Chung Cheng University, Taiwan \\
$^{2}$Posts and Telecommunications Institute of Technology, Ho Chi Minh, Vietnam \\
$^{3}$Université de Bourgogne, Dijon, France \\
$^{4}$Institut de Science Financière et d'Assurances, Lyon, France \\
$^{5}$University of Information Technology, VNU-HCM, Vietnam \\
$^{6}$Posts and Telecommunications Institute of Technology, Hanoi, Vietnam}

\begin{document}
\maketitle

\begin{abstract}
    Semi-Supervised Semantic Segmentation (SSSS) aims to improve segmentation accuracy by leveraging a small set of labeled images alongside a larger pool of unlabeled data. Recent advances primarily focus on pseudo-labeling, consistency regularization, and co-training strategies. However, existing methods struggle to balance global semantic representation with fine-grained local feature extraction. To address this challenge, we propose a novel \textbf{tri-branch} semi-supervised segmentation framework incorporating a \textbf{dual-teacher} strategy, named \textbf{IGL-DT}. Our approach employs SwinUnet for high-level semantic guidance through \textbf{Global Context Learning} and ResUnet for detailed feature refinement via \textbf{Local Regional Learning}. Additionally, a \textbf{Discrepancy Learning} mechanism mitigates over-reliance on a single teacher, promoting adaptive feature learning. Extensive experiments on benchmark datasets demonstrate that our method outperforms state-of-the-art approaches, achieving superior segmentation performance across various data regimes.
\end{abstract}

\section{Introduction}
Semi-Supervised Semantic Segmentation (SSSS) is a critical task in computer vision, aiming to leverage a small set of labeled data alongside a large pool of unlabeled images to enhance segmentation performance. Due to the high cost and labor-intensive nature of pixel-wise annotation \cite{survey}, semi-supervised approaches have gained increasing attention for their ability to mitigate the dependency on extensive labeled datasets while maintaining competitive performance.

To tackle this problem, recent advances in SSSS primarily explore three key strategies: pseudo-labeling \cite{pseudoseg, st++, augseg}, consistency regularization \cite{cct, rc2l, unimatch}, and co-training \cite{cps, ucc, ccvc}. Pseudo-labeling methods generate pseudo-labels for unlabeled samples based on the model’s confident predictions, progressively refining them to improve segmentation performance. For instance, ST++ \cite{st++} enhances self-training by filtering low-confidence predictions and refining pseudo-labels across multiple stages. Similarly, PseudoSeg \cite{pseudoseg} incorporates Self-Attention Grad-CAM (SGC) maps to improve pseudo-label quality and mitigate noise. Meanwhile, consistency regularization enforces prediction stability across different perturbations, enhancing model robustness. CCT \cite{cct} and UniMatch \cite{unimatch} leverage weak-to-strong consistency, while RC2L \cite{rc2l} applies region-based contrastive learning to improve feature discrimination. Finally, co-training approaches enhance generalization by using multi-branch networks to exchange knowledge. CPS \cite{cps} adopts a dual-teacher framework, where two networks supervise each other through cross-generated pseudo-labels, while UCC \cite{ucc} integrates uncertainty estimation to filter unreliable predictions. Additionally, recent approaches often employ Teacher-Student frameworks \cite{mean-teacher, dual-teacher}, where the teacher model generates pseudo-labels to guide the student’s learning process. Those approaches enhance learning by enabling the student model to benefit from the teacher model’s knowledge, which improves generalization and robustness of the student model, especially when labeled data is scarce. 

Despite recent advancements, existing methods often struggle to balance global semantic representation learning with fine-grained local feature extraction. To address this, we introduce a novel tri-branch semi-supervised segmentation framework that incorporates a dual-teacher strategy. Our approach leverages two distinct teacher networks—SwinUnet \cite{swin-unet} and ResUnet \cite{unet}—to guide the student model from complementary perspectives. Specifically, we propose Global Context Learning to transfer high-level semantic knowledge from SwinUnet, Local Regional Learning to refine fine-grained feature representations from ResUnet, and Discrepancy Learning to prevent over-reliance on a single teacher while promoting adaptive feature learning. Unlike conventional Teacher-Student frameworks, which typically employ identical backbones, potentially limiting learning capacity and leading to model collapse when both networks exhibit similar behavior, our framework integrates both a transformer-based backbone (SwinUnet) and a CNN-based backbone (ResUnet). This design enables the student to learn diverse, complementary knowledge, effectively capturing both global and local information. By integrating these three components into a unified semi-supervised learning paradigm, our method enhances both global contextual understanding and local detail preservation, resulting in superior segmentation performance under limited-label scenarios. Our contributions can be summarized as follows: 
\begin{itemize}
    \item This work presents \textit{IGL-DT}, a dual-teacher semi-supervised semantic segmentation framework that effectively integrates global and local feature learning. By leveraging SwinUnet and ResUnet as complementary teachers, the proposed approach balances high-level contextual understanding with fine-grained spatial details.
    \item We propose \textit{Global Context Learning} to transfer global information from the teacher model to the student, while \textit{Local Regional Learning} distills local information. These mechanisms operate iteratively to ensure that the student can learn each objective effectively and benefit from both global and local feature representations.
    \item The Discrepancy Learning module is designed to prevent over-reliance on a single teacher, promoting adaptive feature learning and improving generalization in limited-label scenarios.
    \item The proposed method achieves state-of-the-art segmentation performance on the general benchmark datasets, PASCAL VOC 2012 \cite{pascal} and Cityscapes \cite{cityscapes}.
\end{itemize}

\section{Related Works}

\begin{figure*}[t]
    \centering
    \includegraphics[width=0.9\linewidth]{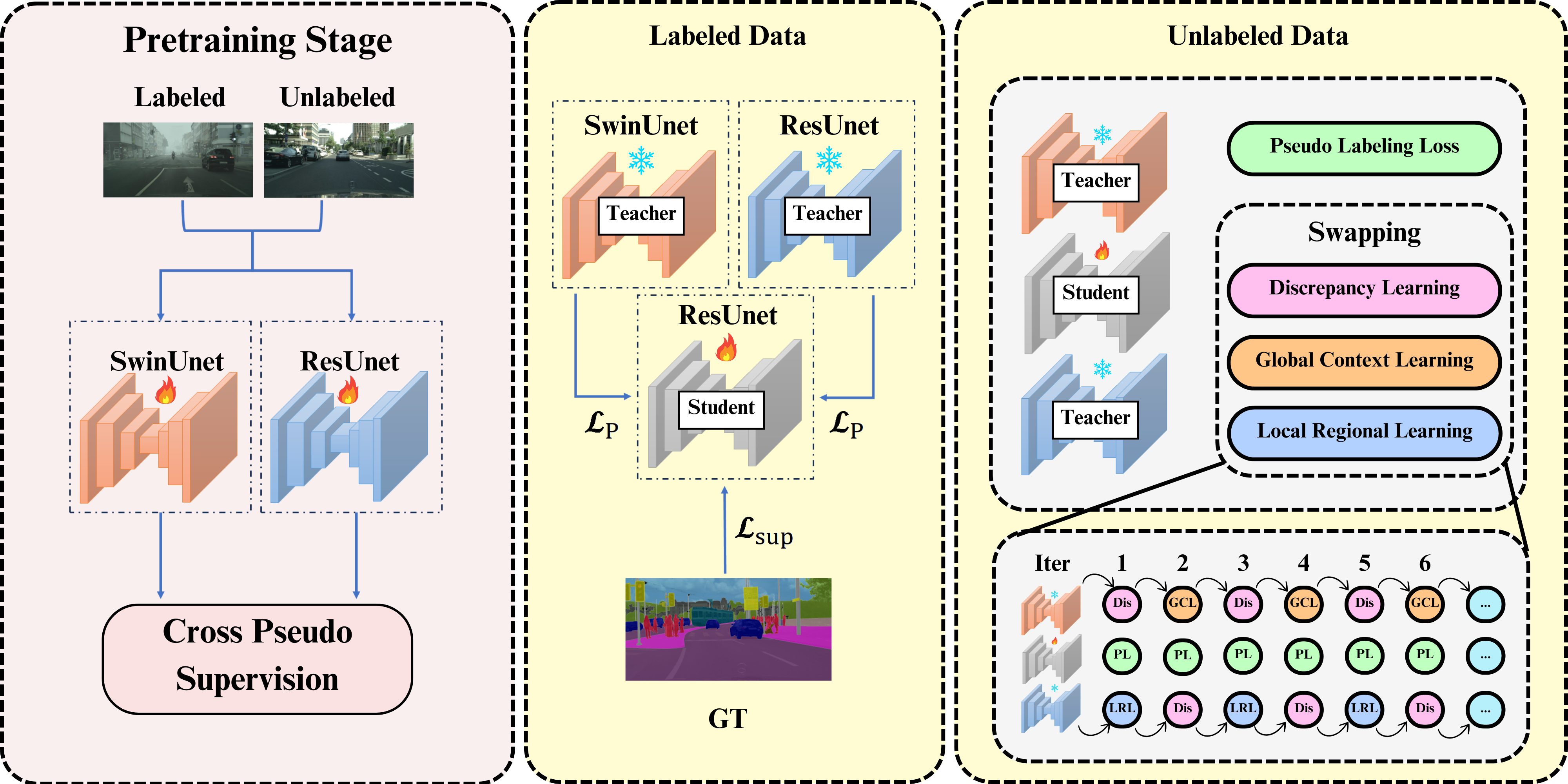}

    \caption{Demonstration of our proposed \textbf{IGL-DT} framework, which consists of two training stages: the \textbf{Teacher Pretraining} Stage (light red color) and the \textbf{Student Learning} Stage (light yellow color). In the \textit{Teacher Pretraining} Stage, the dual-teacher model, initialized with two distinct backbones, is trained using the Cross Pseudo Supervision (CPS) approach to improve weight initialization for generating more reliable pseudo-labels. In the \textit{Student Learning} Stage, the student model learns from labeled data using both pseudo-labels from the teacher models and ground truth labels, optimized through $\mathcal{L}_{sup}$ and $\mathcal{L}_P$. For unlabeled data, the student leverages \textbf{Discrepancy Learning}, \textbf{Pseudo-Labeling Loss}, and an \textbf{iterative swapping} mechanism between \textbf{Global Context Learning} and \textbf{Local Regional Learning}, guided by both teacher models.}
    \label{fig:main-figure}
\end{figure*}

Semi-Supervised Semantic Segmentation (SSSS) is a fundamental task in semi-supervised learning, aiming to enhance segmentation performance by effectively utilizing a limited amount of labeled data alongside a large pool of unlabeled data. To achieve this, recent approaches primarily focus on three key strategies: pseudo-labeling \cite{pseudoseg, st++, nguyen2024blurry}, consistency regularization \cite{cct, unimatch}, and co-training \cite{cps, nguyensemi, ccvc}, all of which contribute to improving model generalization.

Among these, pseudo-labeling plays a crucial role by generating pseudo-labels for unlabeled data based on prior knowledge learned from labeled samples. For instance, PseudoSeg \cite{pseudoseg} enhances semi-supervised semantic segmentation by generating well-calibrated pseudo-labels through a fusion of model predictions and Self-attention Grad-CAM (SGC) maps, which not only improves consistency training but also reduces label noise. Similarly, ST++ \cite{st++} optimizes self-training by refining pseudo-label selection through confidence-based filtering and progressive refinement, thereby mitigating label noise and enhancing model learning with weak-strong augmentation and a multi-stage training strategy. In addition, AugSeg \cite{augseg} highlights the importance of data augmentation in semi-supervised semantic segmentation by incorporating consistency regularization to improve pseudo-label reliability. Enforcing prediction consistency across diverse augmentations strengthens model robustness and minimizes the impact of noisy pseudo-labels.

Another widely used approach in SSSS is consistency regularization, which aims to enforce model invariance under various feature perturbations\cite{nguyen2024fa}. For example, CCT \cite{cct} improves pseudo-label reliability by enforcing consistency between multiple perturbed versions of the same input at different network levels, thereby reducing the impact of noisy labels and enhancing feature learning. Likewise, RC2L \cite{rc2l} integrates region-level contrastive learning with consistency regularization, ensuring that similar regions are pulled closer while dissimilar ones are pushed apart, ultimately strengthening feature discrimination and model robustness. Furthermore, UniMatch \cite{unimatch} refines weak-to-strong consistency by aligning predictions between weakly and strongly augmented images, utilizing confidence thresholding and stronger augmentation strategies to improve pseudo-label quality.

In contrast, co-training approaches leverage multi-branch networks to learn from different perspectives and exchange complementary knowledge, ultimately refining pseudo-label quality and enhancing model robustness. A notable example is CPS \cite{cps}, which introduces a dual-branch training framework where two networks generate pseudo-labels for each other, thereby reinforcing learning through mutual supervision while reducing the impact of noisy labels. Expanding on this idea, UCC \cite{ucc} adopts uncertainty-guided cross-head co-training, where multiple decoder heads generate pseudo-labels for one another, while uncertainty estimation is used to filter unreliable predictions, further improving model stability. Moreover, CCVC \cite{ccvc} enhances consistency training by enforcing agreement between different augmented views while actively detecting and resolving conflicts in predictions, leading to more reliable pseudo-labels and greater robustness against label noise. More advanced methods that combine global and local networks\cite{ngo2024dual, tran2025mlg2net} to perform co-training also achieved potential results across various datasets.

Teacher-student frameworks have also been widely adopted in semi-supervised segmentation, where a teacher model guides a weaker student network by generating pseudo-labels. This approach is not only a pioneer in the SSL field but also in different deep learning fields such as the vision language model \cite{nguyen2024improving,le2023enhancing}. The Mean-Teacher \cite{mean-teacher} is a classic method that applies exponential moving average (EMA) updates to maintain a stable teacher model and enforces consistency regularization between teacher and student predictions. However, static teacher models can limit knowledge transfer. To overcome this, \cite{dual-teacher} introduces a dual-teacher framework where two temporary teachers are alternated across different training stages. Unlike traditional dual-teacher approaches, where both teachers operate simultaneously, STT switches between teachers over time to prevent teacher-student coupling and ensure diverse pseudo-label supervision, ultimately improving segmentation performance.

While these methods have significantly advanced semi-supervised semantic segmentation, they often struggle to balance global semantic understanding with fine-grained local feature extraction. Moreover, relying on a single teacher network may introduce biases, limiting the effectiveness of knowledge transfer. To address these challenges, our work introduces a dual-teacher framework that leverages complementary global and local information reinforced by pseudo-labeling and Discrepancy Learning to enhance segmentation performance in limited-label scenarios.
\section{Methodology}

To address learning ability challenges, we introduce \textbf{IGL-DT}, a tri-branch framework with a dual-teacher strategy to guide the student network from diverse perspectives. Each teacher model is initialized with a distinct backbone: SwinUnet \cite{swin-unet} and ResUnet \cite{unet}. Let $S$ and $T$ represent the student and teacher models, respectively, while $SU$ and $RU$ denote the SwinUnet and ResUnet architectures. Additionally, $\mathcal{E}$ and $\mathcal{D}$ refer to the encoder and decoder of each model. For example, $\mathcal{E}^{RU}_T$ and $\mathcal{D}^{RU}_T$ denote the encoder and decoder of the teacher network initialized with the ResNet \cite{resnet} backbone. Our framework consists of two training stages, called Teacher Pretraining Stage and Student Learning Stage. In the Teacher Pretraining stage, the two distinct teachers are trained using the Cross Pseudo Supervision (CPS) framework \cite{cps}, enabling them to learn mutual knowledge and generate more reliable pseudo-labels for the student model.

In the Student Training stage, the student model is trained using prior knowledge distilled from both teacher models. First, we introduce Global Context Learning, where the ResUnet student learns global semantic knowledge from the SwinUnet teacher. To further enhance learning, we propose Local Regional Learning, which enables the student to effectively capture fine-grained details from images under the guidance of the RU teacher. However, simultaneously learning from both objectives can make the student model difficult to optimize. To mitigate this issue, we introduce Discrepancy Learning, which encourages the student to diverge from one teacher while focusing more on optimizing other objectives. Finally, we summarize how these proposed components are integrated into a semi-supervised scheme by learning from both labeled and unlabeled data. The proposed framework is illustrated in Figure \ref{fig:main-figure}.

\subsection{Global Context Learning}
\begin{figure}[h]
    \centering
    \includegraphics[width=0.9\linewidth]{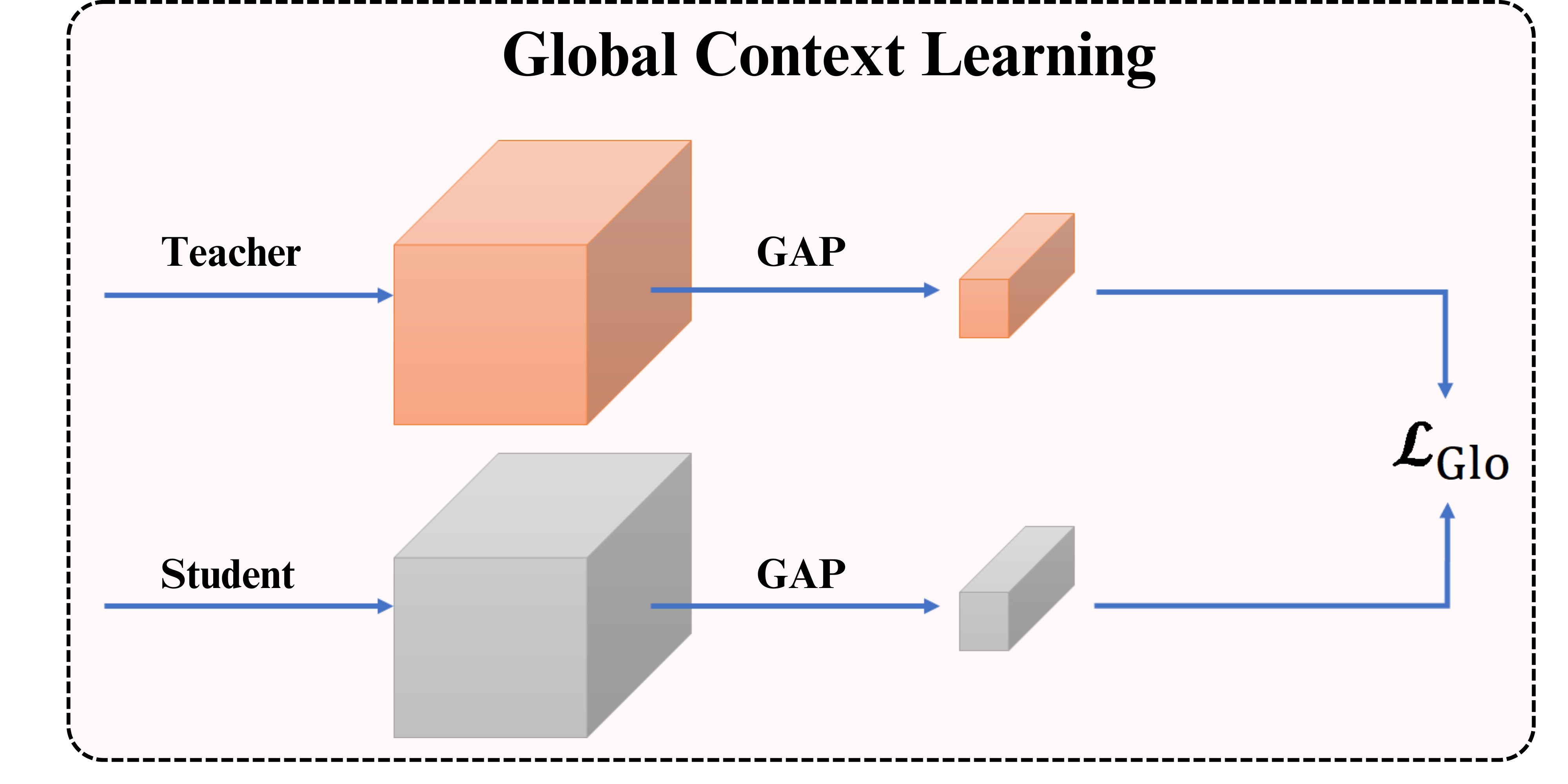}

    \caption{Illustration of the \textbf{Global Context Learning} process in the proposed dual-teacher framework. The teacher and student models extract high-level feature representations, which are processed using Global Average Pooling (GAP) to obtain compact embeddings. The Global Loss ($\mathcal{L}_{Glo}$) enforces consistency between the teacher and student representations, guiding the student to capture global contextual information effectively.}
    \label{fig:global_context_learning}
\end{figure}
To enhance the student network’s ability to capture high-level semantic knowledge, we introduce Global Context Learning. visualized in Figure \ref{fig:global_context_learning}, which enables the student to learn global representations from the SwinUnet teacher. Since SwinUnet effectively models long-range dependencies through self-attention, it serves as an ideal teacher for guiding the student in understanding contextual relationships across the entire image.

Specifically, given an input image, the SwinUnet teacher generates feature maps $F_{SU}$, which encapsulate rich global dependencies. To transfer this knowledge to the student, we extract global context representations by applying Global Average Pooling (GAP) to obtain a compact semantic vector:
\begin{equation}
    K_{SU} = GAP(F_{SU})
\end{equation}

Similarly, the student network $S$ computes its own global representation $K_S$ from its feature maps $F_S$. The goal of Global Context Learning is to align the student's representation with that of the teacher, ensuring that the student effectively captures semantic structures across the entire image.

To enforce this learning process, we introduce $\mathcal{L}_{Glo}$, which measures the discrepancy between the student’s and teacher’s global representations:
\begin{equation}
    \mathcal{L}_{Glo} = \frac{1}{N} \sum_{i=1}^{N} ||K_{S,i} - K_{SU,i}||_1
\end{equation}
where $N$ represents the number of categories, and $|| \cdot ||_1$ denotes the L1 distance. This loss ensures that the student model effectively aligns with the global contextual information extracted from the teacher.

By incorporating Global Context Learning, the student network learns to utilize high-level contextual cues, improving segmentation accuracy while maintaining a lightweight structure. This mechanism plays a crucial role in bridging the gap between high-capacity teachers and compact student models, ultimately enhancing global semantic understanding in semi-supervised learning.

\subsection{Local Regional Learning}
\begin{figure}[t]
    \centering
    \includegraphics[width=1.0\linewidth]{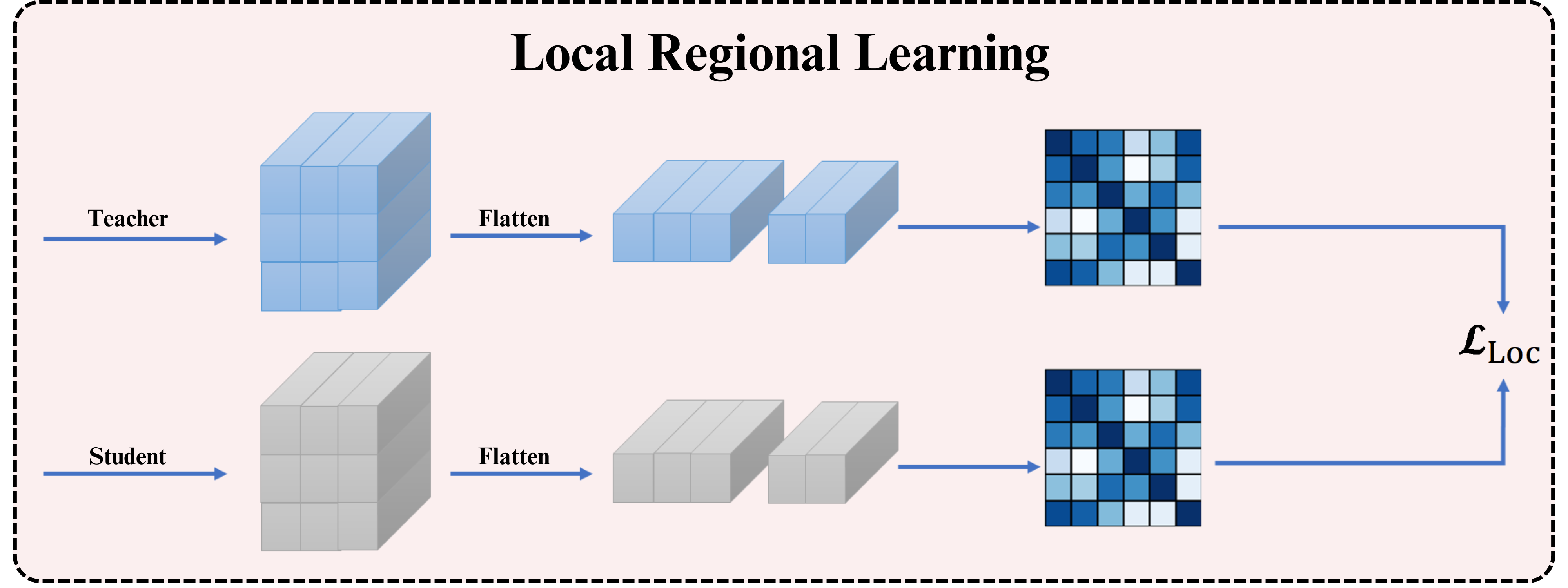}

    \caption{Illustration of the \textbf{Local Regional Learning} process. The teacher and student extract feature maps, which are then flattened into spatial feature representations. The local feature similarities are computed and aligned using the Local Loss ($\mathcal{L}_{Loc}$), encouraging the student to preserve fine-grained spatial details learned by the teacher}
    \label{fig:local_regional_learning}
\end{figure}
To further enhance the student network’s ability to capture fine-grained local structures, we introduce Local Regional Learning. visualized in Figure \ref{fig:local_regional_learning}, which enables the student to learn detailed regional representations from the ResUnet teacher. Since ResUnet excels at preserving spatial details and local context, it serves as an ideal teacher for guiding the student in understanding the relationships between different regions within an image.

Specifically, given an input image, the ResUnet teacher produces feature maps $F_{RU}$, which contain rich local contextual information. To transfer this knowledge to the student, we extract region-wise representations by projecting the feature maps into a set of regional content vectors $V_{RU}$. Each vector represents a localized patch of the image, allowing the student to learn regional correlations.

To ensure effective knowledge transfer, we construct a self-correlation matrix $M_{RU}$ from the teacher’s regional content vectors and similarly compute $M_S$ for the student:
\begin{equation}
    M = sim(V^TV), with\  m_{i,j} = \frac{v^T_i v_j}{||v_i||\ ||v_j||}
\end{equation}
where $M$ encodes the relationship between different regions, and $m_{ij}$ measures the similarity between region $i$ and $j$ using cosine similarity.

To guide the student in effectively capturing local relationships, we introduce $\mathcal{L}_{Loc}$, which minimizes the discrepancy between the student’s and teacher’s regional correlations:
\begin{equation}
    \mathcal{L}_{Loc} = \frac{1}{H_vW_v \times H_vW_v} \sum_{i = 1}^{H_vW_v} \sum_{j = 1}^{H_vW_v} (m_{i,j}^S - m_{i,j}^{RU})^2
\end{equation}
where $H_v W_v$ represents the number of regional content vectors.

By integrating Local Regional Learning, the student network effectively leverages fine-grained regional information, enhancing its ability to preserve object boundaries and local structures. This mechanism ensures that the student model benefits from both global context (via SwinUnet) and local regional details (via ResUnet), leading to improved segmentation performance in a semi-supervised setting.
\subsection{Discrepancy Learning}
While learning from multiple teachers improves the student model's robustness, it can also introduce conflicting guidance, making optimization challenging. To mitigate this, we introduce Discrepancy Learning, which encourages the student to diverge from one teacher while prioritizing optimization on other objectives. This mechanism ensures that the student network does not overly conform to a single teacher, promoting better generalization.

To achieve this, we apply a discrepancy constraint between the student and one selected teacher, enforcing feature separation. Given feature representations extracted from the student ($S$) and a designated teacher ($T$), we define $\mathcal{L}_{Dis}$ as:
\begin{equation}
    \mathcal{L}_{Dis} =  1 + \frac{F_S \cdot F_T}{||F_S||\ ||F_T||}
\end{equation}
where $F_S$ and $F_T$ represent the extracted feature vectors from the student and the chosen teacher (either SwinUnet or ResUnet), respectively. The additional 1 ensures the loss remains non-negative. By minimizing $\mathcal{L}_{dis}$, the student is encouraged to learn distinct representations rather than directly mimicking the selected teacher.

By integrating Discrepancy Learning, we ensure that the student maintains a balance between global knowledge from the SwinUnet teacher and local details from the ResUnet teacher, preventing feature redundancy and improving segmentation performance.

\subsection{Learning with Labeled Data}
To effectively train the student model, we utilize labeled data by enforcing strong supervision at multiple levels. Given an input image $x$ with its ground truth segmentation map $\hat{Y}$, the student network $S$ learns by minimizing the standard supervised segmentation loss. Specifically, we apply a pixel-wise cross-entropy loss to ensure accurate semantic predictions:

\begin{equation} 
    \mathcal{L}_{sup} = \mathcal{L}_{ce}(\hat{Y}, Y) = - \frac{1}{H \times W} \sum_{i=1}^{H \times W}\hat{y}_{i} \log y_{i} 
\end{equation}

where $Y$ represents the predicted probability distribution of the student model, $\hat{Y}$ is the corresponding one-hot encoded ground truth, $H \times W$ denotes the spatial dimensions.

In addition to strong supervision, we enforce consistency between the outputs of the student model and the two teacher models by introducing a consistency loss based on their pseudo-labels. Given pseudo-labels $Y_{SU}$ and $Y_{RU}$ from the SwinUnet ($T_{SU}$) and ResUnet ($T_{RU}$) teachers, respectively, the pseudo-label consistency loss $\mathcal{L}_{P}$ is defined as:

\begin{equation} 
    \mathcal{L}_{P}^l = \mathcal{L}_{ce}(Y, Y_{SU}) + \mathcal{L}_{ce}(Y, Y_{RU}) 
\end{equation}

By leveraging labeled data with strong supervision and consistency loss, we ensure that the student learns accurate semantic representations while benefiting from the complementary knowledge provided by both teachers. The final loss for labeled learning can be written as:
\begin{equation}
    \mathcal{L}_l = \mathcal{L}_{sup} + \mathcal{L}_{P}^l
\end{equation}

\subsection{Learning with Unlabeled Data}
To effectively leverage unlabeled data, we propose two training states, each designed to help the student model focus on optimizing a single objective—Global Context Learning or Local Regional Learning.

In the first state, the student learns global semantic information from the SwinUnet teacher model ($T_{SU}$) through Global Context Learning, while simultaneously applying Discrepancy Learning to the ResUnet teacher model ($T_{RU}$) to encourage focused learning on the current objective and prevent bias toward a single teacher. The loss function for this state is formulated as:

\begin{equation} 
    \mathcal{L}_{s1} = \mathcal{L}_{Glo}(S, T_{SU}) + \alpha\mathcal{L}_{Dis}(S, T_{RU}) 
\end{equation}
where $\alpha$ is a loss weight constraint used to balance the impact of the discrepancy loss on model performance.

In the second state, the student focuses on Local Regional Learning by extracting local contextual information from the ResUnet teacher model ($T_{RU}$) while simultaneously applying Discrepancy Learning to the SwinUnet teacher model ($T_{SU}$). The corresponding loss function is defined as:

\begin{equation} 
    \mathcal{L}_{s2} = \mathcal{L}_{Loc}(S, T_{RU}) + \alpha\mathcal{L}_{Dis}(S, T_{SU}) 
\end{equation}

These two training states alternate iteratively, meaning that if the current iteration follows the first state, the next iteration switches to the second state, and vice versa. This alternating strategy ensures that the student model learns from diverse perspectives, leading to better generalization. The iterative swapping mechanism can be written as:
\begin{equation}
    Swap(\mathcal{L}_{s1}, \mathcal{L}_{s2}) = \frac{1 - (-1)^t}{2} \mathcal{L}_{s1} + \frac{1 + (-1)^t}{2} \mathcal{L}_{s2}
\end{equation}
where \( t \) represents the current iteration index. The term \( (-1)^t \) alternates between \( 1 \) for even \( t \) and \( -1 \) for odd \( t \), ensuring that \(\mathcal{L}_{s1}\) is selected when \( t \) is odd and \(\mathcal{L}_{s2}\) is selected when \( t \) is even.

Additionally, to effectively utilize the complementary information from both teachers, the pseudo-labeling loss is also applied to unlabeled data, which can be formulated as:
\begin{equation}
    \mathcal{L}_P^u = \mathcal{L}_{ce}(Y, Y_{SU}) + \mathcal{L}_{ce}(Y, Y_{RU})
\end{equation}

The final loss function for learning with unsupervised data is defined as:

\begin{equation} 
    \mathcal{L}_{u} = Swap(\mathcal{L}_{s1}, \mathcal{L}_{s2}) + \mathcal{L}_P^u
\end{equation}

\subsection{Overall Objective}
The final objective of our framework can be formulated as:

\begin{equation}
    \mathcal{L} = \mathcal{L}_{l} + \mathcal{L}_{u}
\end{equation}

This loss objective ensures a balanced learning process by incorporating both labeled and unlabeled data. The supervised loss $\mathcal{L}_{l}$ ensures accurate predictions by leveraging ground truth annotations and pseudo-labels from both teachers, while the unsupervised loss $\mathcal{L}_{u}$ enables the model to extract meaningful representations from unlabeled data. By integrating these components, our framework enhances the generalization ability of the student network, allowing it to effectively capture both global semantic information and local contextual details.

\section{Experiments}
\begin{table*}[ht]
    \centering
    \caption{Quantitative comparison of our method with state-of-the-art approaches on the Pascal dataset by showing mIoU scores for different labeled data ratios (1/16, 1/8, and 1/4).}
    \begin{tabular}{p{140pt}|c|c|c|c}
        \hline
        Method & Backbone & 1/16 (662) & 1/8 (1323) & 1/4 (2646)\\
        \hline
        
        Sup-Only & R50 & 62.4 & 68.2 & 72.3\\
        CPS \cite{cps} (CVPR'2021) & R50 & 72.0 & 73.7 & 74.9\\
        PS-MT \cite{ps-mt} (CVPR'2022) & R50 & 72.8 & 75.7 & 76.4\\
        U2PL \cite{u2pl} (CVPR'2022) & R50 & 72.0 & 75.1 & 76.2\\
        ST++ \cite{st++} (CVPR'2022) & R50 & 72.6 & 74.4 & 75.4\\
        CCVC \cite{ccvc} (CVPR'2023) & R50 & 74.5 & 76.1 & 76.4\\
        CorrMatch \cite{corrmatch} (CVPR'2024) & R50 & 75.1 & 76.8 & 77.1\\
        \hline
        \textbf{IGL-DT (Ours)}\cellcolor{gray!15} & R50\cellcolor{gray!15} & \textbf{75.5}\cellcolor{gray!15} & \textbf{77.3}\cellcolor{gray!15} & \textbf{78.0}\cellcolor{gray!15}\\
        \hline
        \hline

        Sup-Only & R101 & 67.5 & 71.1 & 74.2\\
        CPS \cite{cps} (CVPR'2021) & R101 & 74.5 & 76.4 & 77.7\\
        PS-MT \cite{ps-mt} (CVPR'2022) & R101 & 75.5 & 78.2 & 78.7\\
        U2PL \cite{u2pl} (CVPR'2022) & R101 & 74.4 & 77.6 & 78.7\\
        ST++ \cite{st++} (CVPR'2022) & R101 & 74.5 & 76.3 & 76.6\\
        CCVC \cite{ccvc} (CVPR'2023) & R101 & 76.4 & 77.2 & 78.1\\
        CorrMatch \cite{corrmatch} (CVPR'2024) & R101 & 76.9 & 77.7 & 78.4\\
        \hline
        \textbf{IGL-DT (Ours)}\cellcolor{gray!15} & R101\cellcolor{gray!15} & \textbf{77.3}\cellcolor{gray!15} & \textbf{78.2}\cellcolor{gray!15} & \textbf{78.9}\cellcolor{gray!15}\\
        \hline
         
    \end{tabular}
    \label{tab:pascal}
\end{table*}
\subsection{Datasets}
We evaluate our method on two benchmark datasets: PASCAL VOC 2012 \cite{pascal} and Cityscapes \cite{cityscapes}. PASCAL VOC 2012 is a widely used dataset for semantic segmentation, containing 1,464 training images and 1,449 validation images with 21 object classes, including background. Cityscapes is a large-scale urban scene understanding dataset with high-resolution images captured from street scenes across multiple cities. It provides 2,975 training images and 500 validation images, with pixel-level annotations for 19 semantic classes. Due to its complex real-world scenarios and high variability in object appearance, Cityscapes serves as a challenging benchmark for evaluating segmentation models in semi-supervised settings.

\begin{figure*}[t]
    \centering
    \includegraphics[width=1.0\linewidth]{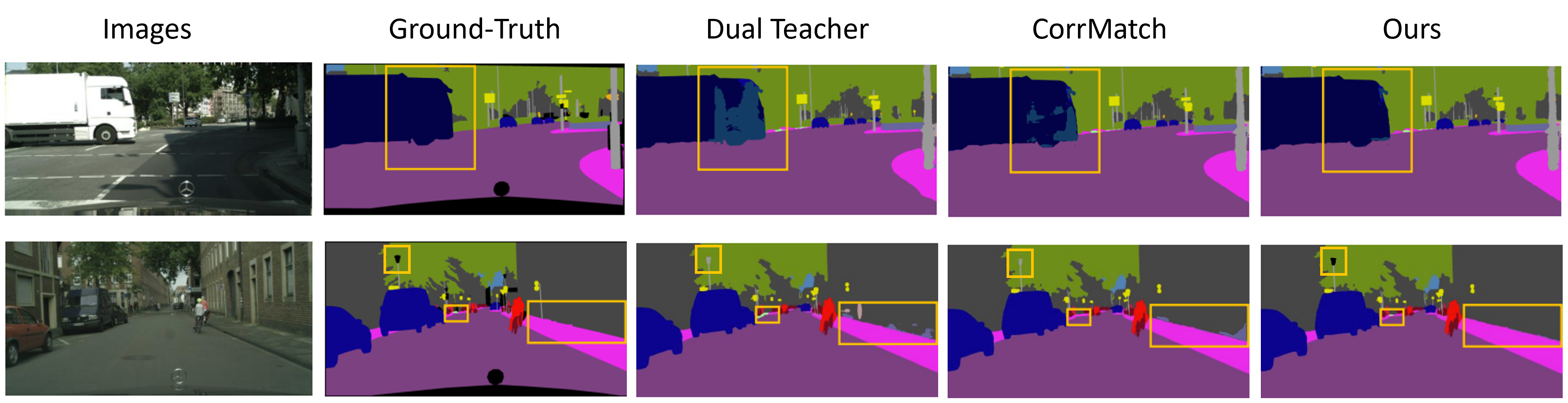}

    \caption{Qualitative comparison of semantic segmentation results on the Cityscapes dataset. The first and second columns show the input images and ground truth annotations. The subsequent columns illustrate different methods' predictions. Our method demonstrates superior performance in preserving fine-grained details and accurately segmenting small or occluded objects (highlighted in yellow boxes)}
    \label{fig:Visualization}
\end{figure*}

\subsection{Implementation Details}
Following previous works, we use DeepLabV3+ \cite{deeplabv3+} as our segmentation model. The local teacher is based on WideResNet \cite{wideresnet}, referred to as ResUnet, while the global teacher utilizes the Swin Transformer \cite{swin}, referred to as SwinUnet. For a fair comparison with previous approaches, the student model backbone is ResNet \cite{resnet}. The hyperparameter $\alpha$ is set to 0.1 for both dataset. In this work, we use the mean Intersection-over-Union (mIoU) as our evaluation metric.

For the PASCAL VOC 2012 dataset, we use WideResNet-50 and WideResNet-101 as the local teacher backbones, responsible for capturing fine-grained spatial features, while SwinUnet serves as the global teacher, modeling long-range dependencies. We follow the standard semi-supervised label partitioning protocol, where the training set is divided into 1/16, 1/8, 1/4, and 1/2 labeled splits, with the remaining data used as unlabeled samples. Optimization is performed using SGD with a momentum of 0.9 and weight decay of 5e-4. The initial learning rate is set to 0.01, following a polynomial decay schedule with a power factor of 0.9. The batch size is set to 24, and the student model is trained for 80 epochs. Input images are cropped to $512\times 512$.

For the Cityscapes dataset, we use WideResNet-101 as the local teacher backbone. We follow the 1/16, 1/8, 1/4, and 1/2 label partitioning protocols, where only a fraction of the 2,975 training images are labeled, and the remaining images are treated as unlabeled data. We employ SGD with a momentum of 0.9 and weight decay of 5e-4. The initial learning rate is set to 0.002, following a polynomial decay schedule. Due to the high resolution of Cityscapes, we set the batch size to 16 (8 labeled + 8 unlabeled), and the model is trained for 240 epochs. Input images are cropped to $512 \times 512$.

\begin{table}[h]
    \centering
    \caption{Quantitative comparison of our method with state-of-the-art approaches on the Cityscapes dataset. Results are reported for different labeled data proportions (1/16, 1/8, 1/4, and 1/2).}
    \begin{tabular}{p{90pt}|c|c|c|c}
        \hline
        Method & 1/16 & 1/8 & 1/4 & 1/2 \\
        \hline

        Sup-Only & 66.3 & 72.8 & 75.0 & 78.0\\
        CPS \cite{cps} & 69.8 & 74.3 & 74.6 & 76.8\\
        PS-MT \cite{ps-mt} & - & 76.9 & 77.6 & 79.1\\
        U2PL \cite{u2pl} & 74.9 & 76.5 & 78.5 & 79.1\\
        UniMatch \cite{unimatch} & 76.6 & 77.9 & 79.2 & 79.5\\
        Dual-Teacher \cite{dual-teacher} & 76.8 & 78.4 & 79.5 & 80.5\\
        CorrMatch \cite{corrmatch} & 76.9 & 78.4 & 79.5 & 80.6\\
        \hline
        \textbf{IGL-DT (Ours)}\cellcolor{gray!15} & \textbf{77.5}\cellcolor{gray!15} & \textbf{78.8}\cellcolor{gray!15} & \textbf{80.0}\cellcolor{gray!15} & \textbf{81.1}\cellcolor{gray!15}\\
        \hline
    \end{tabular}
    \label{tab:cityscapes}
\end{table}  

\subsection{Comparison with state-of-the-art methods}
\textbf{Quantitative Results.} On the PASCAL VOC 2012 dataset, our method achieves state-of-the-art performance across different labeled data ratios, as shown in Table \ref{tab:pascal}. Specifically, with the ResNet-50 (R50) backbone, our approach attains 75.5\%, 77.3\%, and 78.0\% mean Intersection-over-Union (mIoU) for the 1/16 (662 images), 1/8 (1323 images), and 1/4 (2646 images) splits, respectively, surpassing prior methods such as CorrMatch (CVPR 2024) and CCVC (CVPR 2023). Similarly, with the ResNet-101 (R101) backbone, our method achieves 77.3\%, 78.2\%, and 78.9\% mIoU for the same splits, consistently outperforming previous approaches. These results highlight the effectiveness of our dual-teacher framework, demonstrating superior segmentation performance with limited labeled data. On the Cityscapes dataset, our IGL-DT framework achieves superior performance across different labeled data proportions, as presented in Table \ref{tab:cityscapes}. Compared to CorrMatch (CVPR 2024), which attains 76.9\%, 78.4\%, 79.5\%, and 80.6\% mIoU for the 1/16, 1/8, 1/4, and 1/2 splits, respectively, our method consistently outperforms it with 77.5\%, 78.8\%, 80.0\%, and 81.1\%. This improvement demonstrates the effectiveness of integrating both global and local feature learning, reinforcing the robustness of our semi-supervised segmentation framework in urban scene understanding.

\textbf{Qualitative Results.} We present visual comparisons of segmentation results on the Cityscapes dataset to demonstrate the effectiveness of our approach. As illustrated in Figure \ref{fig:Visualization}, our method produces more refined segmentation maps, capturing finer details and reducing misclassifications compared to baseline models. The highlighted regions (yellow boxes) indicate challenging areas where previous methods struggle, such as small objects, object boundaries, and occluded regions. Our approach significantly improves the segmentation quality by leveraging the complementary strengths of local and global teachers, effectively modeling both fine-grained spatial details and long-range dependencies. Notably, our method achieves better delineation of objects like traffic signs, pedestrians, and vehicles, which are often misclassified or omitted in prior approaches.

\subsection{Ablation Studies}

\begin{figure}[h]
    \centering
    \includegraphics[width=1.0\linewidth]{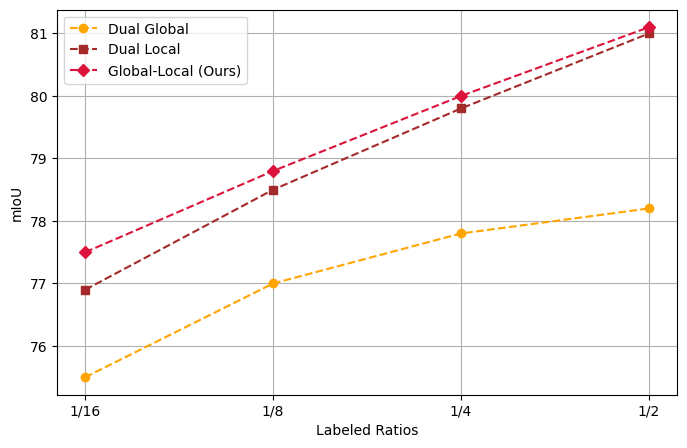}

    \caption{Ablation Studies on Backbone Selection}
    \label{fig:ablation_bs}
\end{figure}

\textbf{Backbone Selection.} Figure \ref{fig:ablation_bs} demonstrates the effectiveness of our Global-Local framework, which integrates knowledge from a Global model (transformer backbone) and a Local model (Wide CNN backbone) to guide the student learner. The Dual-Global and Dual-Local settings represent dual-teacher configurations where both teacher models share the same backbone: SwinUnet for Dual-Global and ResUnet for Dual-Local. As shown in Figure \ref{fig:ablation_bs}, when both teacher models use ResUnet, the student model exhibits only a marginal improvement, suggesting that relying solely on local feature extraction limits segmentation performance. In contrast, using SwinUnet for both teachers leads to a significant boost in accuracy, emphasizing the importance of global contextual learning. However, the best performance is achieved when both backbones are combined in our Global-Local framework, leveraging complementary features from both network architectures. This highlights the advantage of integrating both global semantic understanding and fine-grained local feature refinement in semi-supervised segmentation.

\textbf{Objective Combination}. Table \ref{tab:ablation_oc} presents an ablation study evaluating the impact of different loss components in our framework. We investigate the contributions of Pseudo Labeling Loss ($\mathcal{L}_P^u$) on unlabeled data, Global Context Learning Loss ($\mathcal{L}_{Glo}$), Local Regional Learning Loss ($\mathcal{L}_{Loc}$), and Discrepancy Learning Loss ($\mathcal{L}_{Dis}$). From the results, using only $\mathcal{L}_P^u$ achieves a baseline IoU of 73.8\%. Incorporating Global Context Learning Loss ($\mathcal{L}_{Glo}$) improves performance to 74.2\%, while Local Regional Learning Loss ($\mathcal{L}_{Loc}$) alone slightly decreases performance to 73.2\%, suggesting that global context contributes more significantly to feature alignment. When combining both global and local learning losses, the model achieves 74.4\%, demonstrating their complementary effects. Finally, integrating Discrepancy Learning Loss ($\mathcal{L}_{Dis}$) further enhances performance to 75.5\%, confirming its role in reducing teacher bias and improving student generalization. These results emphasize the importance of jointly leveraging global and local feature learning, while discrepancy learning plays a crucial role in achieving state-of-the-art performance.

\begin{table}[h]
    \centering
    \begin{tabular}{c|c|c|c|c}
        \hline
        $\mathcal{L_P}^u$ & $\mathcal{L}_{Glo}$ & $\mathcal{L}_{Loc}$ & $\mathcal{L}_{Dis}$ & mIoU\\
        \hline
        \checkmark & & & & 73.8\\
        & \checkmark & \checkmark & \checkmark & 74.2\\
        \checkmark & \checkmark & & & 73.2\\
        \checkmark & & \checkmark & & 74.4\\
        \checkmark & \checkmark & \checkmark & & 74.9\\
        \checkmark & \checkmark & \checkmark & \checkmark & \textbf{75.5}\\
        \hline
    \end{tabular}
    \caption{Ablation Studies on Objective Combination}
    \label{tab:ablation_oc}
\end{table}

\section{Conclusion}
In this paper, we introduced a novel tri-branch semi-supervised segmentation framework that effectively integrates Global Context Learning and Local Regional Learning using a dual-teacher strategy. Our approach leverages SwinUnet as the global teacher to capture high-level semantic representations and ResUnet as the local teacher to refine fine-grained details. To further enhance learning, we incorporated Discrepancy Learning, which prevents the student model from overfitting to a single teacher and encourages adaptive feature learning. Extensive experiments on PASCAL VOC 2012 and Cityscapes datasets demonstrate that our method achieves state-of-the-art performance, outperforming prior semi-supervised segmentation approaches. Despite these advancements, our method still has some limitations. The reliance on dual-teacher supervision introduces additional computational costs, which could be optimized in future work.

\section{Acknowledgement}

We thank the support from AI VIETNAM for partially funding and providing the GPU for the numerical calculations.

{
    \small
    \bibliographystyle{ieeenat_fullname}
    \bibliography{main}
}


\end{document}